\documentclass{article}
\usepackage[utf8]{inputenc}
\usepackage{graphicx}
\usepackage[margin=1in]{geometry}
\usepackage{amsmath, amssymb}
\usepackage{url}

\title{Smooth Kolmogorov Arnold networks enabling structural knowledge representation}

\author{
    Moein E. Samadi$^1$, Younes Müller$^1$, Andreas Schuppert$^{*,1}$ \\
    \\
    $^1$Institute for Computational Biomedicine, RWTH Aachen University, Aachen, Germany. \\
    $^*$\textit{Correspondence:} aschuppert@ukaachen.de
}

\date{}

\begin{document}

\maketitle

\begin{abstract}
Kolmogorov-Arnold Networks (KANs) offer an efficient and interpretable alternative to traditional multi-layer perceptron (MLP) architectures due to their finite network topology. However, according to the results of Kolmogorov and Vitushkin, the representation of generic smooth functions by KAN implementations using analytic functions constrained to a finite number of cutoff points cannot be exact. Hence, the convergence of KAN throughout the training process may be limited. This paper explores the relevance of smoothness in KANs, proposing that smooth, structurally informed KANs can achieve equivalence to MLPs in specific function classes. By leveraging inherent structural knowledge, KANs may reduce the data required for training and mitigate the risk of generating hallucinated predictions, thereby enhancing model reliability and performance in computational biomedicine.
\end{abstract}

\subsection*{Introduction}
Recently, Liu et al. \cite{liu2024kan} proposed Kolmogorov-Arnold networks (KANs) as a promising alternative to multi-layer perceptrons (MLPs) for the representation of generic non-linear functions $f$ mapping an $n$-dimensional data space to univariate real outputs: $f \in C^0 (\mathbb{R}^n, \mathbb{R}^1)$.

In contrast to MLP, the Kolmogorov-Arnold representation theorem guarantees that finite KANs with \textit{a priori} defined topology can represent all functions. All continuous functions $f \in C^0 (\mathbb{R}^n, \mathbb{R}^1)$ can be approximated by means of a network consisting of $2n+1$ univariate functional nodes $u_i \in C^0 (\mathbb{R}^1, \mathbb{R}^1)$ in one hidden layer, where the functions $u_i$ are non-linear and nested by linear functional nodes. Hence, networks of this type can conceptually serve as alternatives to MLPs, where non-linearities are hidden in an \textit{a priori} finite set of univariate nodes. Non-linear interaction between the variables is generated by linear intermediate multivariate nodes.

The price to be paid for the universal representation of all functions $f$ by finite networks is the loss of smoothness of the univariate nonlinear functions, even if the overall functions $f$ to be represented are smooth, prohibiting efficient implementations. Hence, the recent results reported by Liu et al. \cite{liu2024kan} showing that deep KAN topologies reduce the numerical challenges arising from the loss of smoothness may pave the route to efficient and interpretable alternatives to the established MLP concepts. However, the interference of smoothness and the representation of generic functions by finite, \textit{a priori} defined networks may be essential for efficient training rates, with implications for future implementations.

Here we will first discuss the role of smoothness in the representation of generic functions using finite networks of nested smooth functions, followed by a discussion of its consequences for generalized KANs, especially for realizing payoffs from the implementation of \textit{a priori} known network structures derived from the real system under consideration.

\subsection*{Smoothness and finite nested functions}

For KANs, Vitushkin proved the existence of even analytic functions $f$ that cannot be represented by KANs using differentiable node functions $u_i$ \cite{vitushkin1964proof}. He showed that Kolmogorov's representation theorem is not valid for smooth functions $f, u_i \in C^k (\mathbb{R}^n, \mathbb{R}^0)$ for all $k \geq 1$ \cite{vitushkin1954hilbert}. He discussed the representation of generic smooth functions $f \in C^k (\mathbb{R}^n, \mathbb{R}^1)$ by finite networks nesting smooth node functions depending on fewer inputs $n'$ than $f$. Any finite network nesting smooth functions $u_i \in C^{k'}(\mathbb{R}^{n'}, \mathbb{R}^1)$ to an overall representation of a smooth function $f \in C^k (\mathbb{R}^n, \mathbb{R}^1)$ \cite{vitushkin1978representation,marchenkov2013interpolation} cannot represent all functions of the type $C^k (\mathbb{R}^n, \mathbb{R}^1)$ if:
\begin{equation}
    \frac{k'}{n'} > \frac{k}{n}
\end{equation}

for all node function $u_i$. Hence, a finite network of KAN type is equivalent to MLP's only if the relation $\frac{k'}{n'} \leq \frac{k}{n}$ holds for at least the essential node functions.

Therefore, in the case of implementations of KANs using nonlinear node functions $u_i$ of lower dimensions ($n' < n$) than the overall dimensionality $n$ of the input-output function $f$, the smoothness $k'$ of the $u_i$ is restricted by the upper bound $k' \leq k \frac{n'}{n}$, resulting in $k' < k$. If the smoothness $k'$ of the inner nodes is implemented such that it does not satisfy the Vitushkin conditions, e.g., by choosing $k' = k$, the equivalence to universal MLPs is no longer guaranteed. 
Therefore, representing even smooth high-dimensional functions f using low-dimensional node functions $u_i$ requires the implementation of highly irregular node functions eventually reducing convergence rates of training.

The local representation of a function $f \in C^k (\mathbb{R}^n, \mathbb{R}^1)$ with arbitrarily high $k$, e.g., an analytic function, by KAN with a given, finite KAN structure requires reproducibility of any derivatives of f by the respective derivatives of the KAN. Suppose the given KAN nests $m$ univariate functions $u_i \in C^k (\mathbb{R}^1, \mathbb{R}^1), i \in \{1, \ldots, m\}$ by means of $c$ linear coupling functions $L_j, j \in \{1, \ldots, c\}$. Each linear coupling function $L_j$ is characterized by a projection vector $w_j$ that nests the outputs of the $u_i$ and the inputs $x$: $w_j \in \mathbb{R}^{m+n}$. In practice, in feed-forward networks, the dimension of the projection vectors $w_j$ will be much less than $m+n$, such that $m+n$ serves as an \textit{a priori} upper bound for the number of parameters to be fit to the data. For the local representation of $f$, all partial derivatives of $f$ must be represented by the KAN at a given input vector $x_0 \in \mathbb{R}^n$. As the KAN is assumed to have a feedforward structure, each $p$-th derivative of the KAN-approximation of $f$ can be represented by a finite set of $p$-th derivatives of the $u_i$ with respect to their inputs according to the chain rule. As the $u_i$ are all univariate and the $L_j$ are linear, the $p$-th derivative of any $u_j$ at a fixed point can be given as $P_u (u_1^{(0)}, \ldots, u_m^{(0)}, \ldots, u_1^{(p)}, \ldots, u_m^{(p)}) \times P_w(w)$, where $P_u$ denotes a polynomial operating on the derivatives of the $u_i$, and $P_w$ denotes a polynomial on all linear parameters $w = \{w_1,...,w_c \}$ of the network, respectively. Apparently, $P_u$ and $P_w$ are determined by the KAN topology. In any case, any local $p$-th derivative of $f$ is represented by a function depending on the $(p + 1) \times m$ values of $\{u^{(0)}, \ldots, u^{(p)}\}(x_0)$ and the $n \times |w| \leq m \times (n+m)$ linear parameters. Hence, all local $p$-th derivatives have to be represented by given polynomials depending on $N_p \leq (p+1)m + m(n+m)$ parameters. The dimension of the vector space of the local $p$-th derivatives:

\begin{equation}
    \left\{ \frac{\partial^p}{\partial x_1 \ldots \partial x_p} f \right\}(x_0)
\end{equation}

which can be represented by the network is limited by $N_p$ for all combinations of partial derivatives. As the number of combinations of $p$-th derivatives grows with $O(p^{n-1})$, for $n \geq 3$ the number of combinations has the lower bound $\binom{n}{2} \frac{p^2}{2}$. Hence, for each KAN with fixed topology, fixing $m$ and $n > 2$, the KAN cannot represent all possible $p$-th local derivatives if 

\begin{equation}
    \left| \left\{ \frac{\partial^p}{\partial x_1 \ldots \partial x_p} f \right\}(x_0) \right| > \binom{n}{2} \frac{p^2}{2} > (p+1)m + m(n+m) \geq N_p.
\end{equation}

As $\binom{n}{2} \frac{p^2}{2}$ grows polynomially with $p$ whereas $N_p$ grows only linearly, for each finite KAN, it exists a limiting level of smoothness which can be represented, limiting the approximation capabilities of the given KAN. However, for any function $f$, there may exist a KAN topology with $m$ nodes that allows a perfect approximation if the topology is chosen according to $f$.

Given that numerical representations typically employ locally smooth functions for approximation almost everywhere, we hypothesize that implementing both the input-output functions $f$ and the node functions $u_i$ within the same function space will impose a $k' = k$ relationship almost everywhere. This, in turn, would result in the non-universality of finite networks of nested functions. Therefore, a perfect approximation of any function by a given finite KAN might be hard to realize (in contrast to a given MLP topology), as the function $f$ may lie outside the space of functions approximated by a given KAN structure. However, in case f cannot be represented precisely by the KAN, but if f is near to the subspace of representable functions, f may be approximated reasonably by the KAN anyway. In that case, the convergence of the training process may be poor in accordance with the partially slow convergence in training reported in \cite{liu2024kan}.

\subsection*{Consequences of smoothness in finite networks of nested functions}

However, for a given topology of finite nested functions, there exist subsets of functions $f \in C^{k'}(\mathbb{R}^{n'}, \mathbb{R}^1)$ that can be represented. In these cases, the higher-order partial derivatives fit the structures of $P_u$ and $P_w$, such that these functions can avoid Vitushkin's smoothness restrictions for generic functions. These subsets of representable functions have been characterized for generic nested functions with tree-structured networks, where the network structure is equivalent to the structure of the system to be represented \cite{fiedler2008local,schuppert2000extrapolability,von2014hybrid}. This result can be used for the implementation of \textit{a priori} known functional structures of the system to be represented in KANs using smooth nodes.

Such structurally informed smooth KANs (which have been named hybrid models) can be trained on significantly reduced data sets and allow extrapolation into sparsely scanned data regions \cite{fiedler2008local}. Both features have been demonstrated in various applications, e.g., in chemical engineering \cite{schuppert2000extrapolability,von2014hybrid}. An implementation using TensorFlow is available at \url{https://github.com/JRC-COMBINE/HybridML}, with a detailed description in \cite{schuppert2011efficient}.

Using the characterization of the relation between network topology and the subset of functions that can be represented for tree-structured networks, the inverse problem for network reconstruction can be solved with implications demonstrated in bioinformatics \cite{merkelbach2022hybridml}. Moreover, an equivalence of finite nested smooth KANs with semi-linear hyperbolic PDE systems has been demonstrated \cite{merkelbach2022hybridml}, allowing the Hadamard initial value problem for such PDE systems to be solved by training appropriately structured smooth KANs from data.

Recently, the concept of smooth KANs has been extended to discrete functions, which play a crucial role in medical data analytics. Although differentiability is no longer defined, efficient training and improved explainability compared to the state of the art have been demonstrated \cite{samadi2024hybrid,samadi2024noisecut,e2022training}. These implementations, however, are based on mathematical proofs valid only for tree-structured functional networks. Although in practice non-tree structures show similar numerical behavior, to our knowledge, a mathematical proof is missing, even though non-tree network structures have been implemented in applications.

\begin{figure}
    \centering
    \includegraphics[scale=0.6]{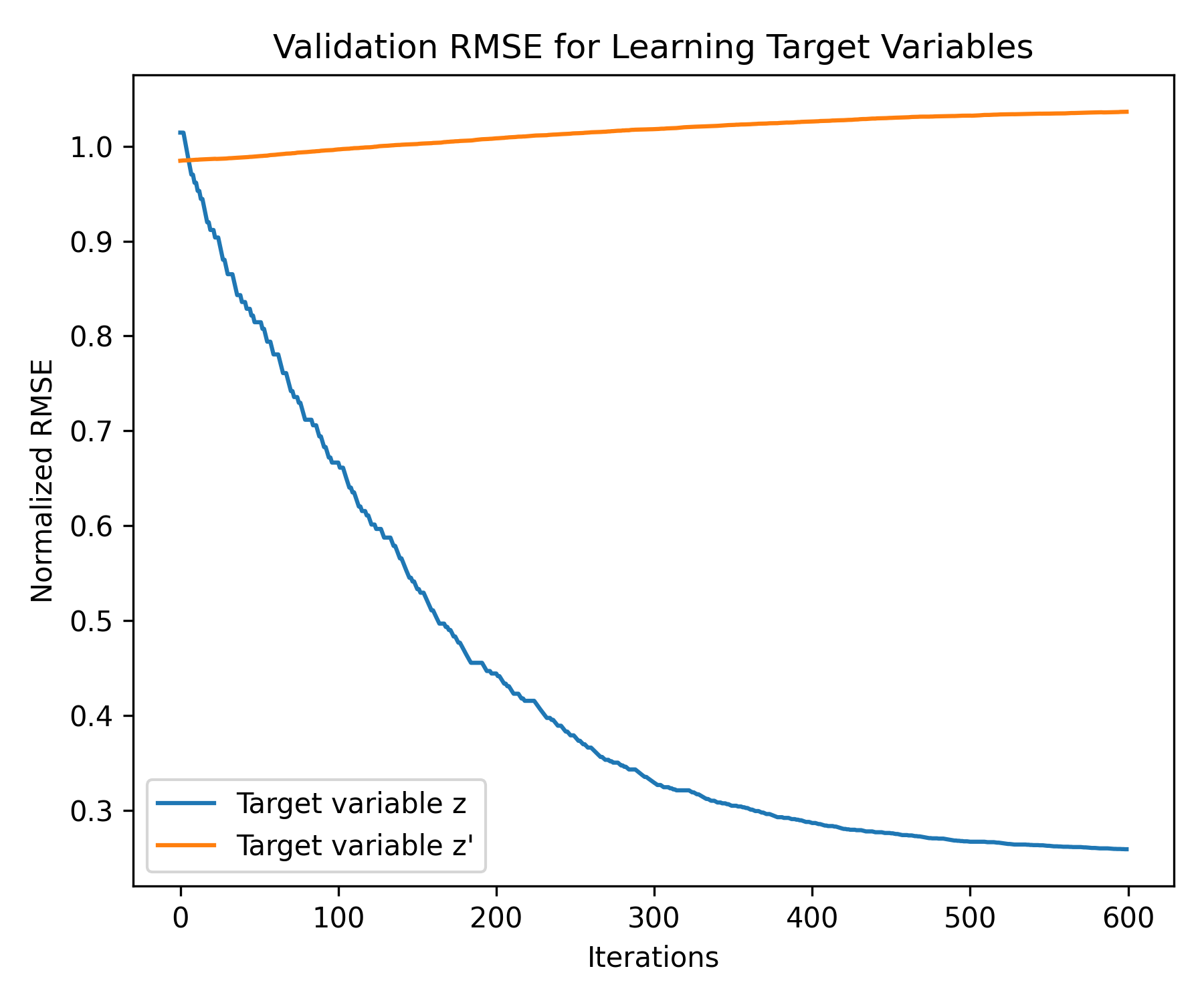}
    \caption{Convergence of the validation RMSE of $w(u(x_1, x_2), v(y_1, y_2))$ for learning the target variables $z = x_1^2 x_2 + y_1 y_2^2$ and $z' = x_1 y_1 y_2 + x_1 x_2 y_2$ by strctured XGBoost regressor model. The model structure is well-suited for predicting $z$, as shown by the decreasing RMSE. 
    However, it struggles to predict $z'$, indicating that $z'$ lies outside the representable function space of the model.}
    \label{fig:1}
\end{figure}

To demonstrate the relationship between adapted structures and the convergence of model training, we designed an experiment to train a nested set of three black-box models (XGBoost regressor models) with a fixed structure to predict target variables $z = x_1^2 x_2 + y_1 y_2^2$ and $z' = x_1 y_1 y_2 + x_1 x_2 y_2$, derived from a feature space in $\mathbb{R}^{4}$. Both functions, $z$ and $z'$, have been represented by the same network structure of nested functions. The structure links black-box models $u(x_1, x_2)$ and $v(y_1, y_2)$, which predict intermediate variables using distinct feature sets, while a third model, $w(u(x_1, x_2), v(y_1, y_2))$, combines these predictions to output the final result. Hence, the function $z$ is adapted to the model structure, whereas $z'$ is outside the representable function space. When training the XGBoost regressor models, we normalized the root mean square error (RMSE) of the decision-making black-box model, $w(u(x_1, x_2), v(y_1, y_2))$, by the standard deviation of the primary target variables, $z$ and $z'$. This training approach effectively minimized the loss for learning the target variable $z$, as evidenced by the convergence of validation RMSE over training iterations in Figure \ref{fig:1}. However, it was ineffective in minimizing the validation RMSE for learning the target variable $z'$.

\subsection*{Conclusion}

As outlined by Liu et al. \cite{liu2024kan}, deep KAN architectures may provide a promising route towards interpretable and efficient implementations of representations of high-dimensional functions. However, as discussed above, the role of smoothness should be taken into account because of the specific, non-intuitive interaction between network topologies and smoothness. Integrating deep KANs with informed network topologies might open new routes towards systematically unraveling intrinsic unknown structures of complex mechanisms with non-linear interactions, as well as enabling data-efficient training and avoiding hallucinated predictions in sparsely sampled data areas. Hence, we expect that in areas like medicine, where interpretability and extrapolation into sparse data ranges are crucial, a network-informed smooth KAN approach will lead to higher acceptance of AI solutions \cite{frohlich2018hype}.


\end{document}